# Stuttgart Open Relay Degradation Dataset (SOReDD)

**Benjamin Maschler[1], Angel Iliev, Thi Thu Huong Pham, Michael Weyrich**

University of Stuttgart, Institute of Industrial Automation and Software Engineering, Pfaffenwaldring 47, 70569 Stuttgart, Germany

## 1. INTRODUCTION

Real-life industrial use cases for machine learning oftentimes involve heterogeneous and dynamic assets, processes and data, resulting in a need to continuously adapt the learning algorithm accordingly [1]. Industrial transfer learning offers to lower the effort of such adaptation by allowing the utilization of previously acquired knowledge in solving new (variants of) tasks [2, 3].

Being data-driven methods, the development of industrial transfer learning algorithms naturally requires appropriate datasets for training. However, open-source datasets suitable for transfer learning training, i.e. spanning different assets, processes and data (variants), are rare [3]. With the Stuttgart Open Relay Degradation Dataset (SOReDD)[2] we want to offer such a dataset. It provides data on the degradation of different electromechanical relays under different operating conditions, allowing for a large number of different transfer scenarios.

Although such relays themselves are usually inexpensive standard components, their failure often leads to the failure of a machine as a whole due to their role as the central power switching element of a machine. The main cost factor in the event of a relay defect is therefore not the relay itself, but the reduced machine availability. It is therefore desirable to predict relay degradation as accurately as possible for specific applications in order to be able to replace relays in good time and avoid unplanned machine downtimes.

Nevertheless, data-driven failure prediction for electromechanical relays faces the challenge that relay degradation behavior is highly dependent on the operating conditions [4], high-resolution measurement data on relay degradation behavior is only collected in rare cases, and such data can then only cover a fraction of the possible operating environments. Relays are thus representative of many other central standard components in automation technology.

## 2. RELAY DEGRADATION

Failure prediction for electromechanical reed relays can be done based on voltage curves, contact resistance, and various other operating parameters. The relays examined for this dataset are switched by a small control current to either block or release a larger load current. The switching operation itself is based on the magnetic influence of a ferromagnetic contact unit induced by the control current [5].

During each switching operation, there is a short time period in which the air gap between the contact tongue and the contact point of the contact tongue is no longer large enough to prevent the voltage from flashing over. An arc briefly occurs, which (in the long term) damages the contact surfaces, resulting in a change in the contact surface

---

[1] Corresponding author, ORCID: 0000-0001-6539-3173, email: benjamin.maschler@ias.uni-stuttgart.de, phone: +49 711 685 67295, fax: +49 711 685 67302

condition with an effect on the contact resistance and which ultimately can lead to welding the contact surfaces together and thus to failure of the relay. The extent of the damage depends, among other things, on the load voltage and load current [5, 6].

## 3. TEST BENCH

In order to document the degradation of electromechanical relays by measuring various operating parameters, a dedicated test bench (see Fig. 1, Fig. 2 and Table 1) was developed at the Institute of Industrial Automation and Software Engineering. It switches different relays under different loads in parallel until they fail. Each relay is controlled by a microcontroller on the input side and connected to a load on the output side. The load circuit is operated by a separate, variable voltage supply.

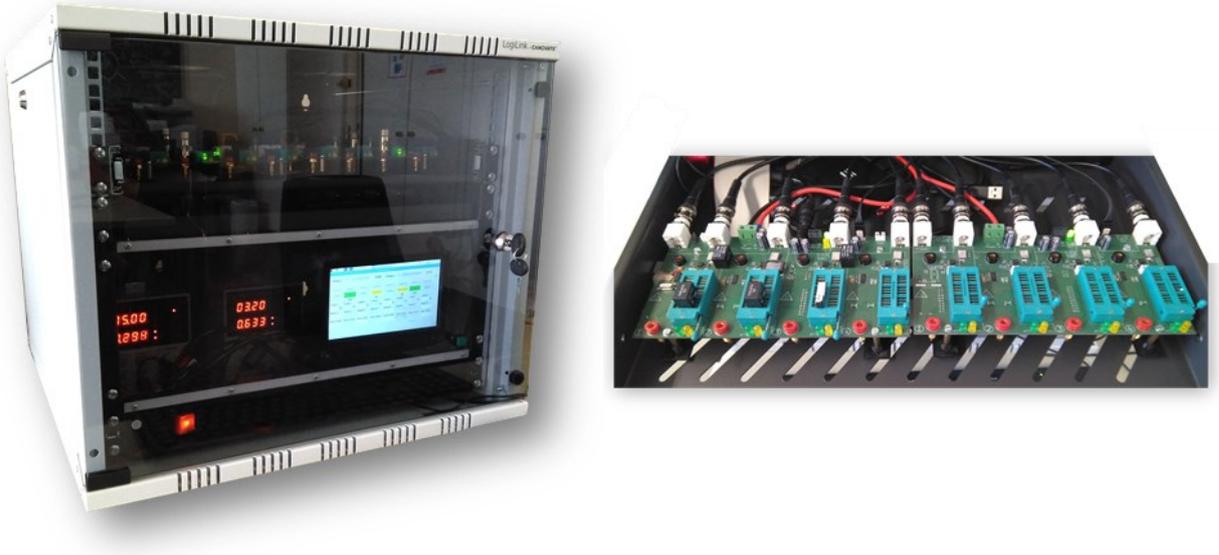

FIGURE 1.    Relay degradation test bench – full view (left) and measuring drawer (right)

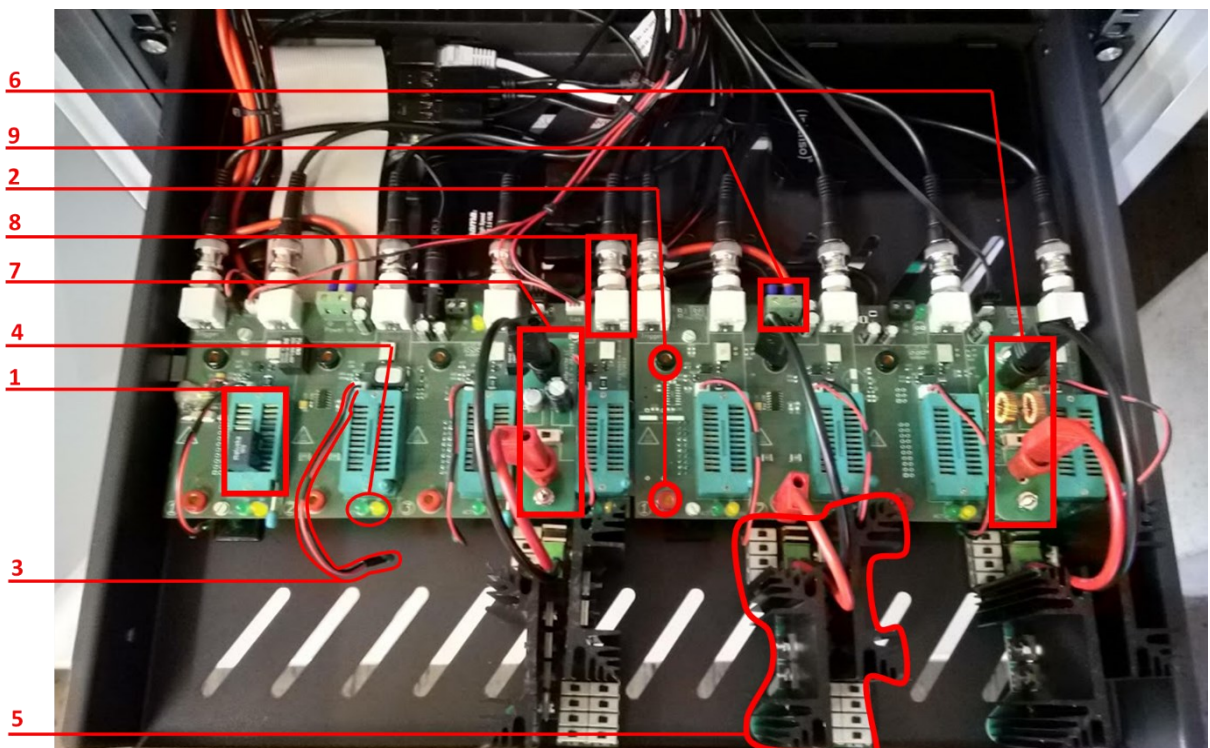

FIGURE 2.    Setup of measuring drawer (see Table 1 for details)

**TABLE 1.** Setup of measuring drawer

| No. in Fig. 2 | Component | Description |
|---|---|---|
| 1 | Relay socket | The slots for the relays are 24-pin ZIF sockets into which different relay types can be plugged. |
| 2 | Load connectors | The connectors for plugging on various loads have the form of jack sockets and are positioned on the left side of each relay socket. |
| 3 | Temperature probe | Behind each relay socket is a connector for a safety temperature probe. |
| 4 | LED | In front of each relay socket are two LEDs. They indicate the operating status of the socket. |
| 5 | Load board R | Ohmic load resistors are introduced into the load circuit by means of a load board R. |
| 6 | Load board L | Inductive load resistors are introduced into the load circuit by means of a load board L (not used in this dataset). |
| 7 | Load board C | Capacitive load resistors are introduced into the load circuit by means of a load board C (not used in this dataset). |
| 8 | Oscilloscope probe connectors | The measuring probes of the USB oscilloscopes are connected to the individual measuring circuits of the two measuring boards via two connector plugs each. |
| 9 | Load power connectors | The adjustable laboratory power supplies are connected to the two measuring boards via two connectors each (Figure 4, item 9). |

The basic circuitry for automated recording of the desired operating parameters of a relay is shown in Fig. 3. The relay is controlled by a microcontroller on the input side and connected to a load on the output side. The load circuit is driven by a separate adjustable laboratory power supply.

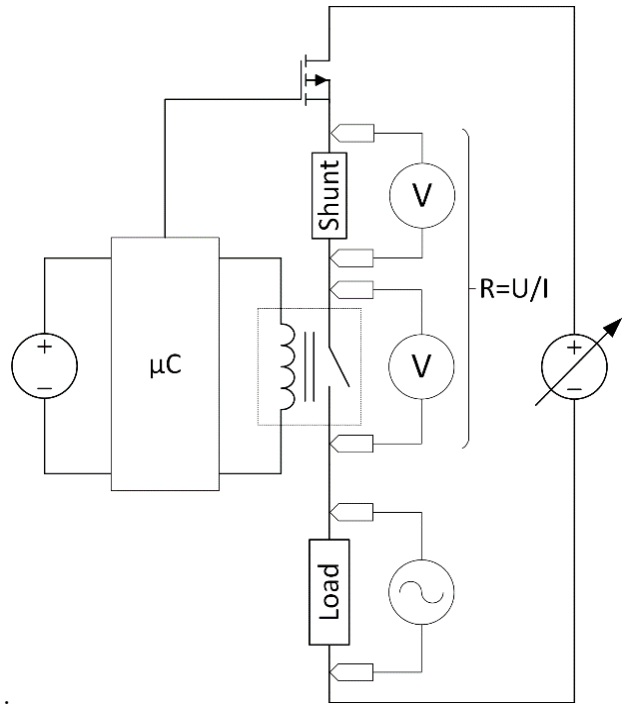

FIGURE 3.   Setup of measuring drawer (see Table 1 for details)

Since the contacts of the relay may not open in the event of a fault, there must be an additional mean of interrupting the load circuit. For this purpose, a transistor is provided in the load circuit, which is controlled by the microcontroller. The output voltage of the relay is measured with an oscilloscope as the voltage across the load. Neglecting the small resistances of the relay switching contact, shunt resistor, and transistor (in the conducting state), the load voltage is exactly equal to the switched output voltage of the power supply, and thus can be referred to as the "output voltage" of the relay. In addition, the measurement of the voltage can also be carried out in the open circuit without a load, in which case it is no longer possible to speak of a voltage across the load. The term "output voltage" is therefore equated in the following with the voltage drop at the load resistance.

The contact resistance of the relay is determined indirectly in the measuring circuit via the voltage drop at the switching contact in the closed state, as well as the current flow through the switching contact. The contact resistance can be calculated from this using Ohm's law. The current flow is measured as a voltage drop across a low-resistance shunt resistor. Thus, only the measurement of two voltages is necessary to determine the contact resistance. The measurement of the voltages can be performed via A/D converters using the existing microcontroller, which means that no external measurement hardware is required

The core of the system is a PC system in the form of a single-board computer, which controls all processes and stores the collected data. It also provides a user interface for interaction with the user via a touch screen. A microcontroller takes over the hardware control of the system. Two remotely controllable laboratory power supplies are used for the power supply of the loads. Laboratory power supplies provide the necessary accuracy and stability of the output voltage for use in the system. All components are housed in a single enclosure. The major components used are listed in Table 2.

**TABLE 2.    Test bench components**

| Component | Type | Datasheet / Manual |
|---|---|---|
| Laboratory Power Supply | Tenma 72-2930 | [7] |
| USB Oscilloscope | Pico Technology Picoscope 3403D | [8] |
| Single-Board-Computer | Raspberry Pi 4B 4GB | [9] |
| Microcontroller | ATmega328P-AU | [10] |

## 4. DATA COLLECTION ROUTINE

Five different types of relays were used in the creation of this dataset. Each was assigned a letter to shorten their respective names without confusing the different types. Table 3 lists the relays and corresponding information.

**TABLE 3.    Relays used**

| Letter | Type | Switching Current [A] | Contact Resistance [Ω] | Closing Time [ms] | Opening Time [ms] | Datasheet / Manual |
|---|---|---|---|---|---|---|
| A | Coto 9007-05-00 | 0.5 | 200 | 0.5 | 0.2 | [11] |
| B | MC3570-1331-051 | 0.5 | 200 | 0.35 | 0.2 | [12] |
| C | Hamlin HE3621A0500 | 0.5 | 150 | 1 | 1 | [13] |
| D | Comus 3570 1331 | 0.5 | 200 | 0.35 | 0.5 | [14] |
| E | TRU Components SIP1A05 | 0.5 | 150 | 1 | 0.5 | [15] |

Using the test bench, 100 relays were opened and closed alternately under 22 different operating conditions defined via the supply voltage and the load resistance (see Table 4) until failure. For every 50th cycle, the voltage curve during opening and closing of the load circuit of the respective relay as well as the contact resistance were measured. Not all of them, but only 32449 samples are included in this dataset. For each relay, the corresponding metadata, i.e. in particular the number of cycles until failure, the supply voltage and the load resistance, were also stored.

**TABLE 4.    Overview of operating conditions**

| Operating condition | Supply voltage [V] | Load resistance [Ω] | Relays | No. of samples | Operating condition | Supply voltage [V] | Load resistance [Ω] | Relays | No. of samples |
|---|---|---|---|---|---|---|---|---|---|
| 1 | 5 | 6 | A1 | 2 | 12 | 20 | 28,5 | A986 | 221 |
| | | | A2 | 310 | | | | A988 | 364 |
| | | | | | | | | A989 | 427 |
| | | | | | | | | A990 | 276 |

| Operating condition | Supply voltage [V] | Load resistance [Ω] | Relays | No. of samples | Operating condition | Supply voltage [V] | Load resistance [Ω] | Relays | No. of samples |
|---|---|---|---|---|---|---|---|---|---|
| | | | | | | | | A991 | 51 |
| | | | | | | | | A992 | 188 |
| 2 | 5 | 3 | B2 | 541 | 13 | 20 | 26,6 | A985 | 76 |
| | | | | | | | | A987 | 476 |
| | | | E1 | 332 | | | | C961 | 90 |
| | | | | | | | | C962 | 649 |
| | | | E2 | 35 | | | | C963 | 527 |
| | | | | | | | | C964 | 507 |
| 3 | 5 | 5 | B5 | 425 | 14 | 20 | 23,5 | B967 | 529 |
| | | | | | | | | B968 | 649 |
| | | | | | | | | B969 | 526 |
| | | | | | | | | B970 | 507 |
| | | | | | | | | B971 | 591 |
| | | | B6 | 670 | | | | B972 | 340 |
| | | | | | | | | B973 | 169 |
| | | | | | | | | B974 | 228 |
| | | | | | | | | C941 | 358 |
| 4 | 5 | 4 | C4 | 298 | 15 | 20 | 20 | B966 | 16 |
| | | | | | | | | B980 | 3 |
| | | | | | | | | B981 | 428 |
| | | | | | | | | B982 | 328 |
| | | | | | | | | B983 | 144 |
| | | | | | | | | D928 | 536 |
| | | | | | | | | D929 | 195 |
| | | | | | | | | D930 | 685 |
| | | | | | | | | D931 | 225 |
| | | | | | | | | D932 | 341 |
| 5 | 10 | 10 | A13 | 472 | 16 | 20 | 21 | B975 | 523 |
| | | | | | | | | B976 | 760 |
| | | | | | | | | B977 | 54 |
| | | | | | | | | B978 | 284 |
| | | | | | | | | B979 | 1 |
| | | | | | | | | D920 | 90 |
| | | | | | | | | D921 | 526 |
| | | | | | | | | D922 | 507 |
| | | | | | | | | D923 | 590 |
| | | | | | | | | D924 | 368 |
| | | | | | | | | D925 | 244 |
| 6 | 10 | 7 | B7 | 29 | 17 | 20 | 30,7 | C945 | 258 |
| | | | C7 | 3 | | | | C946 | 355 |
| | | | C8 | 1 | | | | C947 | 409 |
| 7 | 10 | 7,5 | E7 | 45 | 18 | 20 | 25 | C942 | 578 |

| Operating condition | Supply voltage [V] | Load resistance [Ω] | Relays | No. of samples | Operating condition | Supply voltage [V] | Load resistance [Ω] | Relays | No. of samples |
|---|---|---|---|---|---|---|---|---|---|
| | | | | | | | | C943 | 812 |
| | | | | | | | | C944 | 177 |
| | | | | | | | | C960 | 258 |
| | | | | | | | | C959 | 77 |
| | | | | | | | | C958 | 336 |
| | | | | | | | | C957 | 674 |
| | | | | | | | | C956 | 696 |
| | | | E8 | 7 | | | | C955 | 62 |
| | | | | | | | | C954 | 9 |
| | | | | | | | | C953 | 110 |
| | | | | | | | | C952 | 764 |
| | | | | | | | | C951 | 701 |
| | | | | | | | | C950 | 346 |
| | | | | | | | | C949 | 225 |
| | | | | | | | | C948 | 354 |
| 8 | 10 | 8 | B9 | 601 | 19 | 20 | 24,2 | C940 | 86 |
| | | | B10 | 94 | | | | | |
| 9 | 10 | 8,5 | E9 | 574 | 20 | 20 | 16,6 | D926 | 190 |
| | | | E10 | 19 | | | | D927 | 273 |
| 10 | 10 | 9 | C10 | 78 | 21 | 20 | 20,4 | D917 | 430 |
| | | | | | | | | D918 | 507 |
| | | | | | | | | D919 | 282 |
| 11 | 20 | 32 | A993 | 42 | 22 | 20 | 21,6 | D933 | 44 |
| | | | A994 | 19 | | | | D934 | 58 |
| | | | A995 | 805 | | | | D935 | 696 |
| | | | A996 | 210 | | | | D936 | 503 |
| | | | A997 | 51 | | | | D937 | 2 |
| | | | A998 | 603 | | | | D938 | 805 |
| | | | | | | | | D939 | 210 |

## 5. DATA STRUCTURE

The JSON-files containing the data consist of three name-value-pairs on the topmost level (see Table 5).

**TABLE 5.    Top level of JSON-file structure**

| Name | Type | Description |
|---|---|---|
| metadata | dictionary | The *metadata* dictionary consists of the respective dataset's metadata. |
| contactResistance | dictionary | The *contactResistance* dictionary consists of the respective dataset's contact resistance. |
| switchingCurves | list | The *switchingCurves* list consists of the respective dataset's switching curves, i.e. the load voltage gradient during the switching event. |

The *metadata* dictionary contains the entries listed in Table 6. Two entries, *lastEvent* and *seriesId*, do not exist for some of the datasets (those with a *relayId* in the 900s).

**TABLE 6.** *metadata* structure within JSON-file

| Name | Type | Description |
|------|------|-------------|
| cycles | int | *cycles* indicates the number of cycles before the occurrence of a failure. |
| lastEvent* | string | *lastEvent* indicates the type of failure occurring. |
| numSamples | int | *numSamples* indicates the number of measurements collected for every switching event. *numSamples* divided by *recordingInterval* gives the temporal measurement resolution for the switching curves. |
| recordingInterval | float | *recordingInterval* indicates the duration of measurement collection for every switching event in ms. *numSamples* divided by *recordingInterval* gives the temporal measurement resolution for the switching curves. |
| relayId | int | *relayID* is the unique identifier of every relay of the same type. |
| relayType | string | *relayType* is the unique identifier of the relay type (see Table 3). |
| resistiveLoad | float | *resistiveLoad* indicates the size of the resistive part of the load resistance in Ω. |
| seriesId* | int | *seriesId* is an internal identifier not relevant for use by third parties. |
| storeNth | int | *storeNth* indicates of which (the N-th) cycles measurements are taken. |
| voltage | float | *voltage* indicates the supply voltage in V. |

\* not existent for datasets with *relayId* between 917 and 998

The *contactResistance* dictionary contains the contact resistance measurements for every switching cycle. Table 7 details its internal structure. However, usually not all cycles' data is included in the respective datasets. The contact resistance measurements that are included correspond to the time period where switching curves were collected, too.

**TABLE 7.** *contactResistance* structure within JSON-file

| Level | Name | Type | Description |
|-------|------|------|-------------|
| 1 | name* | string | *name* is "contact". |
| 1 | columns* | list | *columns* contains the two entries "time" and "resistance", labeling the entries of every sublist of *values*. |
| 1 | values | list | *values* contains (unnamed) sublists. |
| 2 | <empty> | list | Each *values* sublist contains two entries, a timestamp** and a contact resistance measurement in mΩ. |

\* not existent for datasets with *relayId* between 917 and 998

\*\* "NaN" for datasets with *relayId* between 917 and 998

The *switchingCurves* list contains the switching curves measurements for *storeNth* switching cycles. Table 8 details its internal structure. However, usually not every *storeNth* cycles' data is included the respective datasets.

**TABLE 8.** *switchingCurves* structure within JSON-file

| Level | Name | Type | Description |
|-------|------|------|-------------|
| 1 | <empty> | dictionary | There is one separate dictionary for every switching curve collected. |
| 2 | name | string | *name* is "waveform". |
| 2 | tags | dictionary | *tags* contains information (*cycle*, *state*) on the respective switching curve. |
| 3 | cycle | int | *cycle* indicates to which switching cycle the switching curve belongs. |
| 3 | state | string | *state* indicates whether the switching event was an "opening" or "closing" event. |
| 2 | columns** | list | *columns* contains the two entries "time" and "voltage", labeling the entries of every sub-list of *values*. |
| 2 | values | list | *values* contains *numSamples* (unnamed) sub-lists. |
| 3 | <empty> | list | Each *values* sub-list contains two entries, a timestamp** and a load voltage measurement in mV. |

\* not existent for datasets with *relayId* between 917 and 998

\*\* "NaN" for datasets with *relayId* between 917 and 998

Fig. 4 depicts excerpts of exemplary switching curves of opening and closing events of the same relay at different cycle numbers, one at the beginning (red), in the middle (green) and at the end (blue) of the relay's lifecycle.

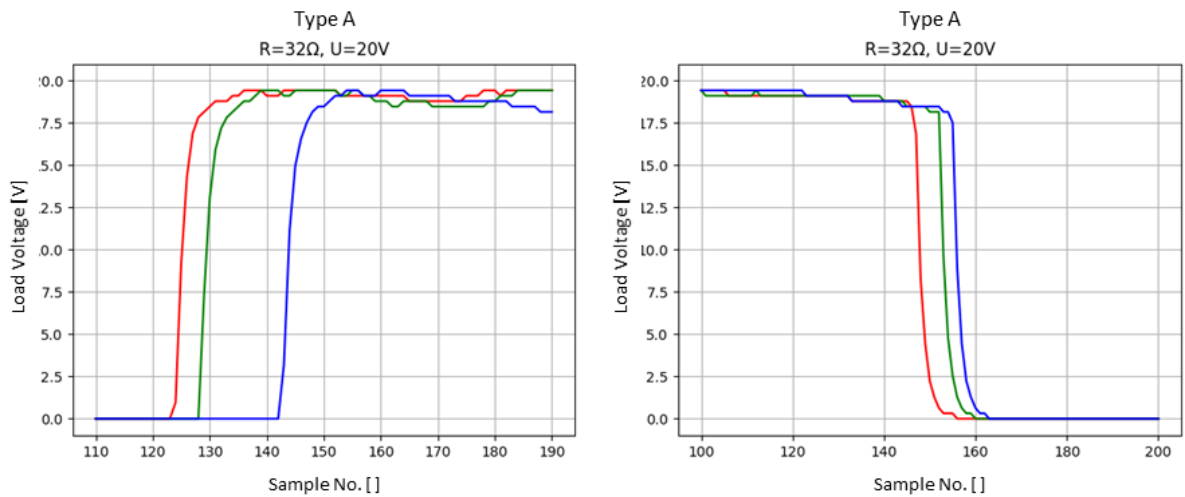

FIGURE 4. **Example switching curves – closing (left) and opening (right) the load circuit**

## 6. ACKNOWLEDGEMENTS

We thank David Kellner, Andre Michelberger and Severin Davis for their contributions to the creation of the test bench.

## REFERENCES

[1] B. Maschler, T. Knodel and M. Weyrich, "Towards Deep Industrial Transfer Learning: Clustering for Transfer Case Selection," 27th IEEE Conference on Emerging Technologies and Factory Automation (ETFA) 2022, Stuttgart, Germany, 2022 (accepted).

[2] B. Maschler, and M. Weyrich, "Deep Transfer Learning for Industrial Automation," IEEE Industrial Electronics Magazine 15 (2), p. 65-75, 2021.

[3] B. Maschler, H. Vietz, H. Tercan, C. Bitter, T. Meisen and M. Weyrich, "Insights and Example Use Cases on Industrial Transfer Learning," 55th CIRP Conference on Manufacturing Systems (CMS) 2022, Lugano, Switzerland, 2022 (accepted).

[4] Y. Xuerong, M. Yue, M. Hang and Z. Guofu, "Degradation failure model of electromagnetic relay," 26th IEEE International Conference on Electrical Contacts (ICEC) 2012, Beijing, China, 2012.

[5] V. Gurevich, "Electric relays: Principles and applications," Taylor & Francis, Boca Raton, USA, 2006.

[6] C.H. Leung and A. Lee, "Contact erosion in automotive DC relays," IEEE Transactions on Components, Hybrids and Manufacturing Technology, 14 (1), p. 101-108, 1991.

[7] Farnell, "Digital-Control and Programmable DC Power Supply User Manual", available at http://www.farnell.com/datasheets/2054525.pdf, last visited on 04.02.2022

[8] Pico Technology, "PicoScope 3000-Serie Datasheet" available at https://www.picotech.com/download/datasheets/PicoScope3000DDMSOSeriesDataSheet-de.pdf, last visited on 04.02.2022

[9] Raspberry Pi Foundation, "Raspberry Pi 4 Model B Datasheet" available at https://datasheets.raspberrypi.com/rpi4/raspberry-pi-4-datasheet.pdf, last visited on 04.02.2022

[10] Atmel, "ATmega328P Datasheet" available at https://ww1.microchip.com/downloads/en/DeviceDoc/Atmel-7810-Automotive-Microcontrollers-ATmega328P_Datasheet.pdf, last visited on 04.02.2022

[11] Coto Technology, "9007 Series/Spartan SIP Reed Relays Datasheet" available at https://www.farnell.com/datasheets/1853229.pdf, last visited on 04.02.2022

[12] Multicomp, "Instrumentation Grade SIP Reed Relay" available at https://www.farnell.com/datasheets/2646307.pdf, last visited on 04.02.2022

[13] Hamlin, "HE3600 Miniature S.I.L. Relay Features and Benefits" available at https://www.farnell.com/datasheets/1633362.pdf, last visited on 04.02.2022

[14] COMUS Relays, "General Purpose SIP Product Data Sheet" available at https://comus-intl.com/wp-content/uploads/2015/01/3570-1331.pdf, last visited on 04.02.2022

[15] TRU Components, "Reed Relay 5V/DC (2270598) Data Sheet" available at https://asset.conrad.com/media10/add/160267/c1/-/en/002270598DS00/datenblatt-2270598-tru-components-sip1a05-reed-relais-1-schliesser-5-vdc-05-a-10-w-sip-4.pdf, last visited on 04.02.2022